%% file: ms.tex
\documentclass[10pt,twocolumn,letterpaper]{article}

\usepackage{iccv}
\usepackage{times}
\usepackage{epsfig}
\usepackage{graphicx}
\usepackage{amsmath}
\usepackage{amssymb}

\usepackage{color}
\usepackage{multicol}
\usepackage{booktabs}
\usepackage{makecell}
\usepackage{pifont}
\usepackage{array,multirow}

\definecolor{darkgreen}{rgb}{0.0, 0.5, 0.0}
\definecolor{brickred}{rgb}{0.8, 0.25, 0.33}
\definecolor{darkblue}{rgb}{0.2,0.2,0.8}
\newcommand{\best}[1]{\underline{\textbf{#1}}}
\newcommand{\sbest}[1]{\textbf{{#1}}}

\usepackage[pagebackref=true,breaklinks=true,letterpaper=true,colorlinks,bookmarks=false]{hyperref}

\iccvfinalcopy 

\def\iccvPaperID{2312} 

\ificcvfinal\fi
\begin{document}

\title{ViCo: Word Embeddings from Visual Co-occurrences}

\author{Tanmay Gupta \qquad Alexander Schwing \qquad Derek Hoiem\\
University of Illinois at Urbana Champaign\\
{\tt\small \{tgupta6, aschwing, dhoiem\}@illinois.edu \url{http://tanmaygupta.info/vico/}}
}

\maketitle
\ificcvfinal\thispagestyle{empty}\fi

\begin{abstract}
We propose to learn word embeddings from visual co-occurrences. Two words co-occur visually if both words apply to the same image or image region. Specifically, we extract four types of visual co-occurrences between object and attribute words from large-scale, textually-annotated visual databases like VisualGenome and ImageNet. We then train a multi-task log-bilinear model that compactly encodes word ``meanings'' represented by each co-occurrence type into a single visual word-vector. Through unsupervised clustering, supervised partitioning, and a zero-shot-like generalization analysis we show that our word embeddings complement text-only embeddings like GloVe by better representing similarities and differences between visual concepts that are difficult to obtain from text corpora alone. We further evaluate our embeddings on five downstream applications, four of which are vision-language tasks. Augmenting GloVe with our embeddings yields gains on all tasks.  We also find that random embeddings perform comparably to learned embeddings on all supervised vision-language tasks, contrary to conventional wisdom.


\end{abstract}

\input{figures/teaser.tex}
\input{sections/intro.tex}

\input{sections/related_work.tex}
\input{sections/model.tex}

\input{sections/visual_cooccur.tex}
\input{sections/analysis.tex}
\input{sections/tasks.tex}

{\small
\bibliographystyle{ieee_fullname}
\bibliography{references}
}


\end{document}

%% file: figures/teaser.tex
\begin{figure}[t]
\begin{center}
\includegraphics[width=\linewidth]{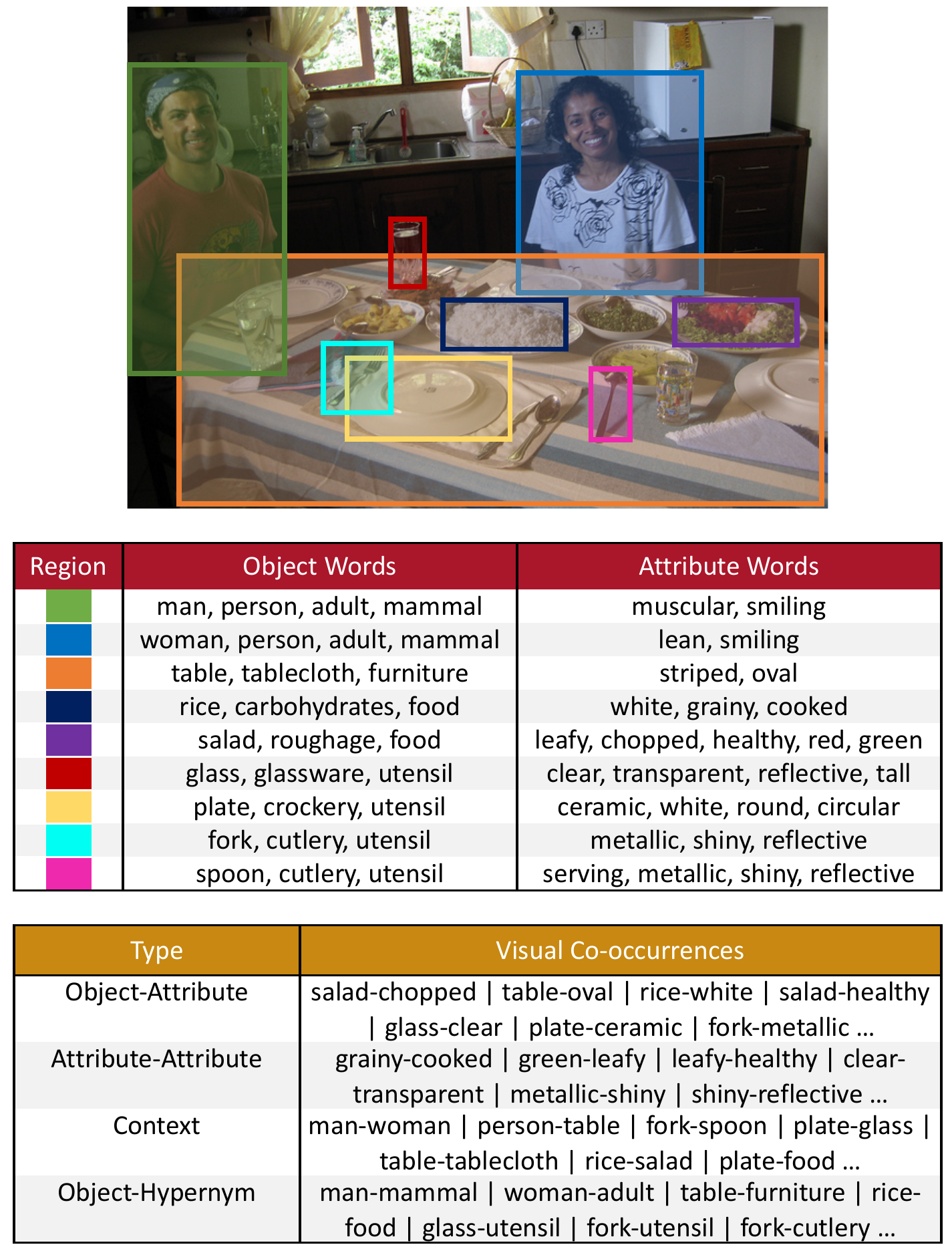}
\end{center}
\vspace{-0.75em}
\caption{\textbf{Visual co-occurrences are a rich source of information for learning word meanings.} The figure shows regions annotated with words and attributes in an image, and the four types of visual co-occurrences used for learning ViCo embeddings.}
\label{fig:teaser}
\vspace{-1em}
\end{figure}

%% file: sections/intro.tex
\section{Introduction}
Word embeddings, \ie, compact vector representations of words, 
are an integral component in many 
language~\cite{seo2017bidaf,he2017deepsrl,lee2017coref,peters2017seqtag,parikh2016ADA,stanovsky2018openie,rashkin2018event2mind} and vision-language models~\cite{massiceti2018flipdial,wang2019imgtxt,wang2018zeroshot,anne2017localizing,plummer2017phraseloc,plummer2018CITE,vasileva2018fashion,gupta2017svlr,shih2016wtl,das2017dialog,jiang2018pythia,karpathy2015captioning,luo2017refexp}. These word embeddings, \eg, GloVe and word2vec, are typically learned from large-scale text corpora by modeling textual co-occurrences. However, text often consists of interpretations of concepts or events rather than a description of visual appearance. This limits the ability of text-only word embeddings to represent visual concepts.

To address this shortcoming, we propose to 
gather co-occurrence statistics of words based on images 
and learn word embeddings from these visual co-occurrences. Concretely, two words co-occur visually if both words are applicable to the same image or image region. 
We use four types of co-occurrences as shown in Fig.~\ref{fig:teaser}: (1) \emph{Object-Attribute} co-occurrence between an object in an image region and the region's attributes; (2) \emph{Attribute-Attribute} co-occurrence of a region; (3) \emph{Context} co-occurrence which captures joint object appearance in the same image; and (4) \emph{Object-Hypernym} co-occurrence between a visual category and its hypernym (super-class).

Ideally, for reliable visual co-occurrence modeling of a sufficiently large vocabulary (a vocabulary size of $400$K is typical for text-only embeddings),  a dataset with all applicable vocabulary words annotated for each region in an image is required. 
While no visual dataset exists with such exhaustive annotations (many  non-annotated words may still be applicable to an image region), large scale datasets like VisualGenome~\cite{kingma2014adam} and ImageNet~\cite{deng2009imagenet} along with their WordNet~\cite{miller1995wordnet} \emph{synset} annotations provide a good starting point. We use ImageNet annotations augmented with WordNet hypernyms to compute Object-Hypernym co-occurrences while the remaining types of co-occurrence are computed from VisualGenome's object and attribute annotations.


To learn ViCo, \ie,  word embeddings from \textbf{Vi}sual \textbf{Co}-occurrences, 
we could concatenate GloVe-like embeddings trained separately for each co-occurrence type via a log-bilinear model. However, in this  na\"ive approach, the dimensionality of the learned embeddings scales linearly with the number of co-occurrence types. To avoid this linear scaling, 
we extend the log-bilinear model by formulating  a \emph{multi-task} problem, 
where learning embeddings from each co-occurrence type constitutes a different task with compact trainable embeddings shared among all tasks. In this formulation the embedding dimension can be chosen independently of the number of co-occurrence types. 

To test ViCo's ability to capture similarities and differences between visual concepts, we analyze performance in an \emph{unsupervised clustering}, \emph{supervised partitioning} (see supplementary material), and a \emph{zero-shot-like} visual generalization setting. The clustering analysis is performed on a set of most frequent words in VisualGenome which we manually label with \emph{coarse} and \emph{fine-grained} visual categories. For the \emph{zero-shot-like} setting, we use CIFAR-100 with different splits of the 100 categories into seen and unseen  sets. In both cases, ViCo augmented GloVe outperforms GloVe, random vectors, \emph{vis-w2v}, or their combinations. Through a qualitative analogy question answering evaluation, we also find ViCo embedding space to better capture relations between visual concepts than GloVe.

We also evaluate ViCo on five downstream tasks -- a discriminative attributes task, and four vision-language tasks. The latter includes Caption-Image Retrieval, VQA, Referring Expression Comprehension, and Image Captioning. Systems using ViCo outperform those using GloVe for almost all tasks and metrics. 
While learned embeddings are typically believed to be important for vision-language tasks, somewhat surprisingly, we find random embeddings compete tightly with learned embeddings on all vision-language tasks. This suggests that either by nature of the tasks, model design, or simply training on large datasets, the current state-of-the-art vision-language models do not  benefit much from learned embeddings. Random embeddings perform significantly worse than learned embeddings in our clustering, partitioning, and zero-shot analysis, as well as the discriminative attributes task, which does not involve images.


To summarize our contributions: (1) We develop a multi-task method to learn a word embedding from multiple types of co-occurrences; 
(2) We show that the embeddings learned from multiple visual co-occurrences, when combined with GloVe, outperform GloVe alone in unsupervised clustering, supervised partitioning, and zero-shot-like analysis, as well as on multiple vision-language tasks; 
(3) We find that performance of supervised vision-language models is relatively insensitive to word embeddings, with even random embeddings leading to nearly the same performance as learned embeddings. To the best of our knowledge, our study provides the first empirical evidence of this unintuitive behavior for multiple vision-language tasks.


%% file: sections/related_work.tex
\section{Related Work}
Here we describe non-associative, associative, and the most recent contextual models of word representation. 

\noindent\textbf{Non-Associative Models.} Semantic Differential (SD)~\cite{osgood1957sd} is among the earliest attempts to obtain vector representations of words. SD relies on human ratings of words on 50 scales between bipolar adjectives, such as `happy-sad' or `slow-fast.' Osgood \etal~\cite{osgood1957sd} further reduced the 50 scales to 3 orthogonal factors. However, the scales were often vague (\eg, is the word `coffee' `slow' or `fast') and provided a limited representation of the word meaning. Another approach involved acquiring word similarity annotations followed by applying Multidimensional Scaling (MDS)~\cite{kruskal1964mds} to obtain low dimensional (typically 2-4) embeddings and then identifying meaningful clusters or interpretable dimensions~\cite{ross1999foodmds}. Like SD, the MDS approach lacked representation power, and embeddings and their interpretations varied based on words (\eg, food names~\cite{ross1999foodmds}, animals~\cite{rips1973semdist}, \etc) to which MDS was applied. 

\noindent\textbf{Associative Models.} The hypothesis underlying associative models is that word-meaning may be derived by modeling a word's association with all other words. Early attempts involved factorization of word-document~\cite{deerwester1990lsa} or word-word~\cite{lund1996hal} co-occurrence matrices. Since raw co-occurrence counts can span several orders of magnitude, transformations of the co-occurrence matrix based on Positive Pointwise Mutual Information (PPMI)~\cite{bullinaria2007ppmi} and Hellinger distance~\cite{lebret2014hellinger} have been proposed. Recent neural approaches like the Continuous Bag-of-Words (CBOW) and the Skip-Gram models~\cite{mikolov2013w2vcorr,mikolov2013w2vnaacl,mikolov2013w2vnips} learn from co-occurrences in local context windows as opposed to global co-occurrence statistics. Unlike global matrix factorization, local context window based approaches use co-occurrence statistics rather inefficiently because of the requirement of scanning context windows in a corpus during training but performed better on word-analogy tasks. Levy~\etal~\cite{levy2014implicitmf} later showed that Skip-Gram 
with negative-sampling performs implicit matrix factorization of a PMI word-context matrix. 

\input{figures/models.tex}

Our work is most closely related to GloVe~\cite{pennington2014glove} which combines the efficiency of global matrix factorization approaches with the performance obtained from modelling local context. We extend GloVe's log-bilinear model to simultaneously learn from multiple types of co-occurrences. We also demonstrate that visual datasets annotated with words are a rich source of co-occurrence information that complements the representations learned from text corpora alone.

\noindent\textbf{Visual Word Embeddings.} There is some work on incorporating image representations into word embeddings. \emph{vis-w2v}~\cite{kottur2016visw2v} uses abstract (synthetic) scenes to learn visual relatedness. The scenes are clustered and cluster membership is used as a surrogate label in a CBOW framework. Abstract scenes have the advantage of providing good semantic features for free but are limited in their ability to match the richness and diversity of natural scenes.  However, natural scenes present the challenge of extracting good semantic features. Our approach uses natural scenes but bypasses image feature extraction by only using co-occurrences of annotated words. ViEW~\cite{hasegawa2017incorporatingvfeat} is another approach to visually enhance existing word embeddings. An autoencoder is trained on pre-trained word embeddings while matching intermediate representations to visual features extracted from a convolutional network trained on ImageNet. ViEW is also limited by the requirement of good image features.    


\noindent\textbf{Contextual Models.} Embeddings discussed so far represent individual words. However, many language understanding applications demand representations of words in context (\eg, in a phrase or sentence) which in turn requires to learn how to combine word or character level representations of neighboring words or characters. The past year has seen several advances in contextualized word representations through pre-training on language models such as ELMo~\cite{peters2018elmo}, OpenAI GPT~\cite{radford2018gpt}, and BERT~\cite{devlin2018bert}. However, building mechanisms for representing context is orthogonal to our goal of improving representations of individual words (which may be used as input to these models). 


%% file: figures/models.tex
\begin{figure*}[t]
\vspace{-0.75em}
\begin{center}
\includegraphics[width=\linewidth]{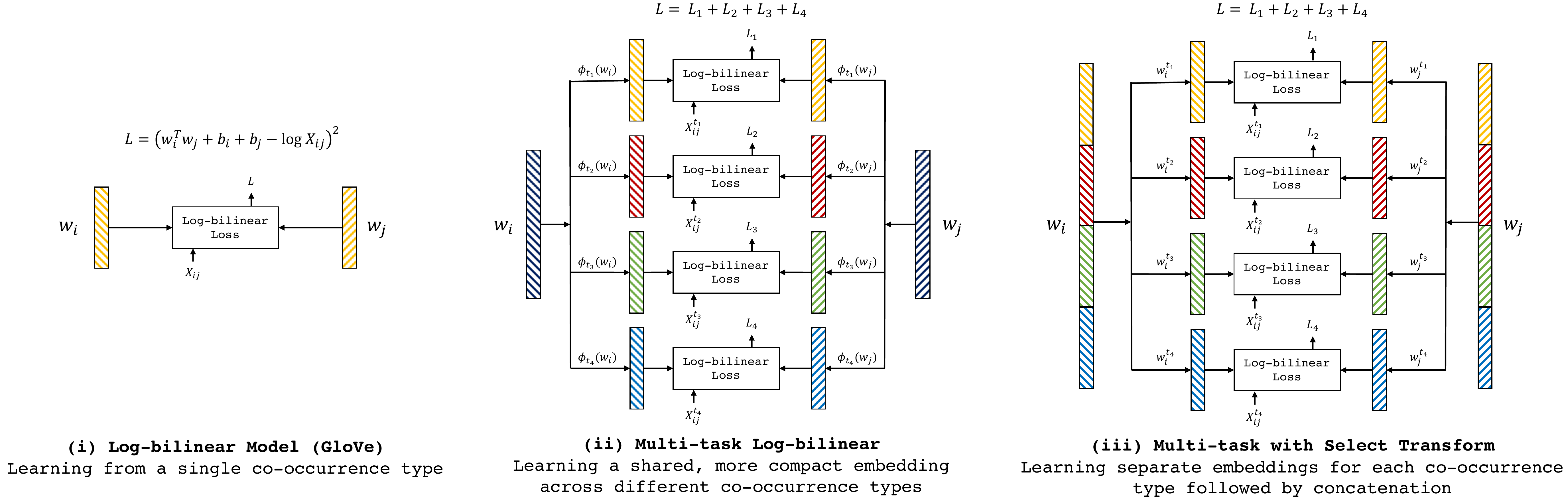}
\end{center}
\vspace{-0.75em}
\caption{\textbf{Log-bilinear models and our multi-task extension.} We show loss computation of different approaches for learning word embeddings $w_i$ and $w_j$ for words $i$ and $j$. The embeddings are denoted by colored vertical bars. (i) shows GloVe's log-bilinear model. (ii) is our multi-task extension to learn from multiple co-occurrence matrices. Word embeddings $w_i$ and $w_j$ are projected into a dedicated space for each co-occurrence type $t$ through transformation $\phi_t$. Log-bilinear losses are computed in the projected embedding spaces. (iii) shows an approach where the different colored regions of $w_i$ (or $w_j$) are allocated to learn from different co-occurrence types. This approach, equivalent to training separate embeddings followed by concatenation, can be implemented in our multi-task formulation using a \emph{select} transform (Tab.~\ref{tab:transforms}). Tab.~\ref{tab:analysis_variants} shows that an appropriate choice of $\phi$ (\eg, \emph{linear}) in the multi-task framework leads to more compact embeddings than (iii) without sacrificing performance since  the correlation between different co-occurrence types is utilized.}
\label{fig:models}
\vspace{-1em}
\end{figure*}

%% file: sections/model.tex
\section{Learning ViCo}
We describe the  GloVe formulation for learning embeddings from a single co-occurrence matrix in Sec.~\ref{sec:glove} and introduce our multi-task extension to learn embeddings jointly from multiple co-occurrence matrices in Sec.~\ref{sec:multitask}. Sec.~\ref{sec:cooccur} describes how co-occurrence count matrices are computed for each of the four co-occurrence types. 

\subsection{GloVe: Log-bilinear Model}~\label{sec:glove}
Let $X_{ij}$ denote the co-occurrence count between words $i$ and $j$ in a text corpus. Also let $\mathcal{N}$ be the list of word pairs with non-zero co-occurrences. GloVe learns $d$-dimensional embeddings $w_i \in \mathbb{R}^d$ for all words $i$ by optimizing 
\begin{equation}~\label{eq:glove}
   \min_{w, b} \sum_{(i,j) \in \mathcal{N}} f(X_{ij})(w_i^Tw_j + b_i + b_j - \log{X_{ij}})^2,
\end{equation}
where $f: \mathbb{R} \rightarrow \mathbb{R}$ is a weighting function that assigns lower weight to less frequent, noisy co-occurrences and $b_i$ is a learnable bias term for word $i$. 

Intuitively, the program in Eq.~\eqref{eq:glove} learns word embeddings such that for any word pair with non-zero co-occurrence, the dot product $w_i^Tw_j$ approximates the log co-occurrence count up to an additive constant. The word meaning is derived by simultaneously modeling the degrees of association of a single word with a large number of other words~\cite{murphy2004concepts}. We also refer the reader to~\cite{pennington2014glove} for more details. 

Note the slight difference between the objective in Eq.~\eqref{eq:glove} and the original GloVe objective: GloVe replaces $w_j$ and $b_j$ with $\tilde{w}_j$ (context vector) and $\tilde{b}_j$ which are also trainable. The GloVe vectors are obtained by averaging $w_i$ and $\tilde{w}_i$. However, as also noted in~\cite{pennington2014glove}, given the symmetry in the objective, both vectors should ideally be identical. We did not observe a significant change in performance when using separate word and context vectors.

\subsection{Multi-task Log-bilinear Model}~\label{sec:multitask}
We now extend the log-bilinear model described above to jointly learn embeddings from multiple co-occurrence count matrices $X^t$, where $t\in{\cal T}$ refers to a type from the set of types ${\cal T}$. 
Also let $\mathcal{N}_t$ and $\mathcal{Z}_t$ be the list of word pairs with non-zero and zero co-occurrences of type $t$ respectively. We learn ViCo embeddings $w_i \in \mathbb{R}^d$ for all words $i$ by minimizing the following loss function 
\begin{multline}~\label{eq:multi_sense_logbilinear}
\hspace{-1.5em}\sum_{t\in \mathcal{T}} 
\sum_{(i,j) \in \mathcal{N}_t} 
(\phi_t(w_i)^T\phi_t(w_j) + b_i^t + b_j^t - \log{X_{ij}^t})^2 \; + \\
\sum_{t\in \mathcal{T}} 
\sum_{(i',j') \in \mathcal{Z}_t}
\max(0, \phi_t(w_{i'})^T\phi_t(w_{j'}) + b_{i'}^t + b_{j'}^t).
\end{multline}
Here $\phi_t: \mathbb{R}^d \rightarrow \mathbb{R}^{d_t}$ is a co-occurrence type-specific transformation function that maps ViCo embeddings to a type-specialized embedding space. $b_i^t$ is a learned bias term for word $i$ and type $t$. We set function $f(X)$ in Eq.~\eqref{eq:glove} to the constant $1$ for all $X$. 
Next, we discuss the transformations $\phi_t$, benefits of capturing different types of co-occurrences, use of the second term in Eq.~\eqref{eq:multi_sense_logbilinear}, and training details. Fig.~\ref{fig:models} illustrates (i) GloVe and versions of our model (ii,iii). \\


\input{tables/transforms.tex}
\input{figures/sim.tex}
\noindent\textbf{Transformations $\phi_t$.} To understand the role of the transformations $\phi_t$ in learning from multiple co-occurrence matrices, consider the na\"ive approach of concatenating $|\mathcal{T}|$ $d_t$-dimensional word embeddings learned separately for each type $t$ using Eq.~\eqref{eq:glove}. Such an approach would yield an embedding with  $d \geq |\mathcal{T}|\min_{t}d_t$ dimensions. For instance, 4 co-occurrence types, each producing embeddings of size $d_t=50$, leads to $d=200$ dimensional final embeddings. Thus, a natural question arises -- \emph{Is it possible to learn a more compact representation by utilizing the correlations between different co-occurrence types?} 

Eq.~\eqref{eq:multi_sense_logbilinear} is  
a multi-task learning formulation where learning from each type of co-occurrence constitutes a different task. Hence, $\phi_t$ is equivalent to a task-specific head that projects the shared word embedding $w\in \mathbb{R}^d$ to a type-specialized embedding space $\phi_t(w)\in \mathbb{R}^{d_t}$. A log-bilinear model equivalent to Eq.~\eqref{eq:glove} is then applied for each co-occurrence type in the corresponding specialized embedding space. We learn the embeddings $w$ and parameters of $\phi_t$  simultaneously for all $t$ in an end-to-end manner. 

With this multi-task formulation   the dimensions of $w$ can be chosen independently of $|{\cal T}|$ or $d_t$. Also note that the new formulation encompasses the na\"ive approach which is implemented in this framework by setting $d=\sum_td_t$, and $\phi_t$ as a slicing operation that `selects' $d_t$ non-overlapping indices allocated for type $t$. In our experiments, we evaluate this na\"ive approach and refer to it as the \emph{select} transformation. We also assess \emph{linear} transformations of different dimensions as described in Tab.~\ref{tab:transforms}. We find that 100 dimensional ViCo embeddings learned with \emph{linear} transform achieve the best performance \vs compactness trade-off. \\




\noindent\textbf{Role of $\max$ term.} Optimizing only the first term given in Eq.~\eqref{eq:multi_sense_logbilinear} can lead to accidentally embedding a word pair from $\mathcal{Z}_t$ (zero co-occurrences) close together (high dot product). To suppress such spurious similarities, we  include the $\max$ term which encourages all word pairs $(i',j') \in \mathcal{Z}_t$ to have a small predicted log co-occurrence 
\begin{equation}
\log{\tilde{X}_{i'j'}^t} = \phi_t(w_{i'})^T\phi_t(w_{j'}) + b_{i'}^t + b_{j'}^t.
\end{equation}
In particular, the second term in the objective linearly penalizes positive predicted log co-occurences of word-pairs that do not co-occur. \\

\input{tables/stats.tex}

\noindent\textbf{Training details.} Pennington~\etal~\cite{pennington2014glove} report Adagrad to work best for GloVe. We found that Adam leads to faster initial convergence. 
However, fine-tuning with Adagrad further decreases the loss. For both optimizers, we use a learning rate of $0.01$, a batch size of $1000$ word pairs sampled from $\mathcal{N}_t$ and $\mathcal{Z}_t$ each for all $t$, and no weight decay. \\

\noindent\textbf{Multiple notions of relatedness.} Learning from multiple co-occurrence types leads to a richer sense of relatedness between words. 
Fig.~\ref{fig:sim} shows that the relationship between two words may be better understood through similarities in multiple embedding spaces than just one. For example, `window' and `door' are related because they occur in context in scenes, `hair' and `blonde' are related through an object-attribute relation, `crouch' and `squat' are related because both attributes apply to similar objects, \etc.

%% file: tables/transforms.tex
\setlength{\tabcolsep}{4pt}
\begin{table}[t]
\centering
\small
\begin{tabular}{l|c|c|c}
\textbf{Transforms} & \boldmath$d$ & \boldmath$d_t$ & \boldmath$\phi_t$\\
\hline
\emph{select (200)} & $200$ & $50 \; \forall \; t$ &  \makecell{$\phi_t(w)=[w[i_0^t],\cdots,w[i_{49}^t]]$ \\ where $\{i_0^t,\cdots,i_{49}^t\}$ are indices \\ pre-allocated for $t$ in $\{0,\cdots,200\}$} \\
\hline
\emph{linear (50)} & $50$ & $50 \; \forall \; t$ & \makecell{$\phi_t(w)=A_tw$ \\ where $A_t\in \mathbb{R}^{50\times50}$} \\
\hline
\emph{linear (100)} & $100$ & $50 \; \forall \; t$ & \makecell{$\phi_t(w)=A_tw$ \\ where $A_t\in \mathbb{R}^{50\times100}$} \\
\hline
\emph{linear (200)} & $200$ & $50 \; \forall \; t$ & \makecell{$\phi_t(w)=A_tw$ \\ where $A_t\in \mathbb{R}^{50\times200}$}
\end{tabular}
\vspace{0.5em}
\caption{\textbf{Description and parametrization of transforms.} ${\phi_t: \mathbb{R}^d \rightarrow \mathbb{R}^{d_t}}$ is a transform for co-occurrence type $t\in \mathcal{T}$. \mbox{\emph{select}} corresponds to approach (iii) in Fig.~\ref{fig:models} that concatenates separately trained $d_t$ dimensional embeddings.}
\label{tab:transforms}
\vspace{-1em}
\end{table}

%% file: figures/sim.tex
\begin{figure}[t]
\begin{center}
\hspace{-1.5em}\includegraphics[width=\linewidth]{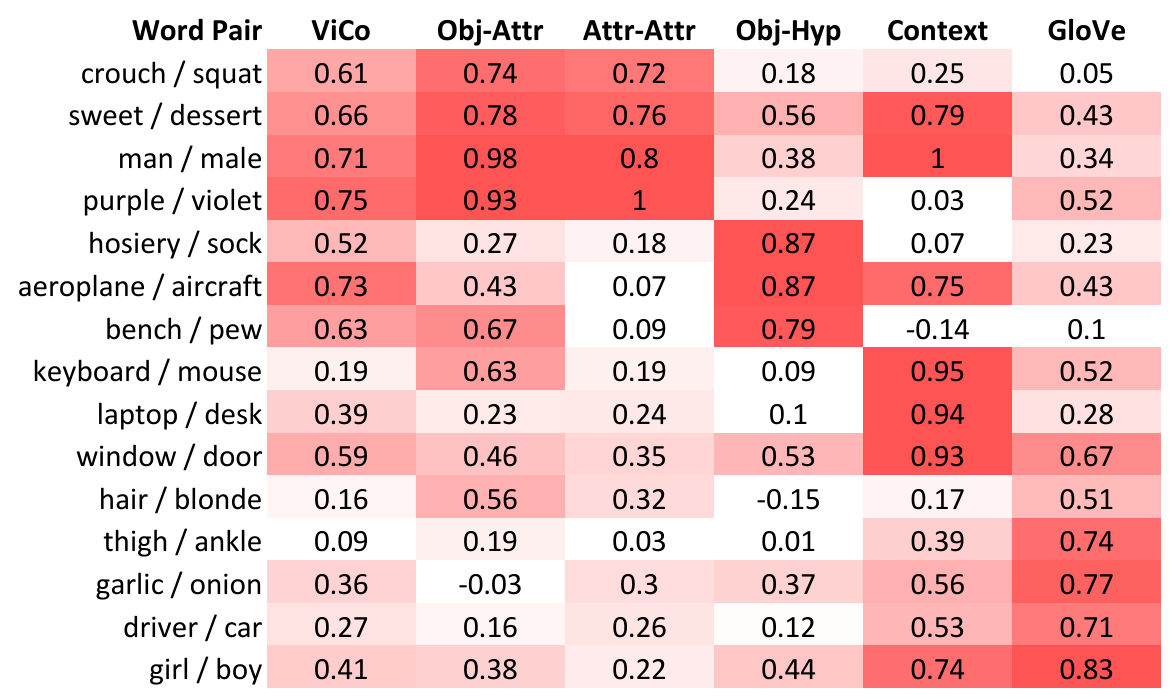}
\end{center}
\vspace{-0.75em}
\caption{\textbf{Rich sense of relatedness through multiple co-occurrences.} Different notions of word relatedness exist but current word embeddings do not provide a way to disentangle those. Since ViCo is learned from multiple types of co-occurrences with dedicated embedding spaces for each (obtained through transformations $\phi_t$), it can provide a richer sense of relatedness. The figure shows cosine similarities computed in GloVe, ViCo(linear) and embedding spaces dedicated to different co-occurrence types (components of ViCo(select)). For example, `hosiery' and `sock' are related through an object-hypernym relation but not related through object-attribute or a contextual relation. `laptop' and `desk' on the other hand are related through context.}
\label{fig:sim}
\vspace{-1em}
\end{figure}

%% file: tables/stats.tex
\renewcommand{\arraystretch}{1.5}
\setlength{\tabcolsep}{2.8pt}
\begin{table}[t]
\centering
\footnotesize
\begin{tabular}{l|cccc|c}
 & \textbf{Obj-Attr} & \textbf{Attr-Attr} & \textbf{Obj-Hyp} & \textbf{Context} & \textbf{Overall}\\
\hline
Unique Words & $15,548$ & $11,893$ & $11,981$ & $25,451$ & $35,476$ \\
\hline
\makecell{Non-zero entries \\ (in millions)} & $1.37$ & $1.37$ & $0.61$ & $8.12$ & $11.48$ \\ 
\end{tabular}
\vspace{0.5em}
\caption{\textbf{Co-occurrence statistics} showing the number of words and millions of non-zero entries in each co-occurrence matrix. For reference, GloVe uses a vocabulary of $400,000$ words with 8-40 billion non-zero entries.}
\label{tab:stats}
\vspace{-1em}
\end{table}
\renewcommand{\arraystretch}{1.0}

%% file: sections/visual_cooccur.tex
\subsection{Computing Visual Co-occurrence Counts}~\label{sec:cooccur}

To learn meaningful word embeddings from visual co-occurrences, 
reliable co-occurrence count estimates are crucial. We use Visual Genome and ImageNet for estimating visual co-occurrence counts. Specifically, we use object and attribute \emph{synset} (set of words with the same meaning) annotations in VisualGenome to get \emph{Object-Attribute} ($oa$), \emph{Attribute-Attribute} ($aa$), and \emph{Context} ($c$) co-occurrence counts. ImageNet \emph{synsets} and their ancestors in WordNet are used to compute \emph{Object-Hypernym} ($oh$) counts. Tab.~\ref{tab:stats} shows the number of unique words and non-zero entries in each co-occurrence matrix. 

Let $\mathcal{T}=\{oa,aa,c,oh\}$ denote the set of four co-occurrence types and $X_{ij}^t$ denote the number of co-occurrences of type $t\in\mathcal{T}$ between words $i$ and $j$. We denote a \emph{synset} and its associated set of words as $\mathcal{S}$. All co-occurrences are initialized to $0$. We now describe how each co-occurrence matrix $X^t$ 
is computed.

\begin{itemize}
    \itemsep 0em
    \item Let $\mathcal{O}$ and $\mathcal{A}$ be the sets of object and attribute synsets annotated for an image region. For each region in VisualGenome, we increment $X_{ij}^{oa}$ by $1$, for each word pair $(i,j)\in \mathcal{S}_o\times\mathcal{S}_a$, and for all \emph{synset} pairs $(\mathcal{S}_o,\mathcal{S}_a) \in \mathcal{O}\times\mathcal{A}$. $X_{ji}^{oa}$ is also incremented unless $i=j$.
    \item For each region in VisualGenome, we increment $X_{ij}^{aa}$ by $1$, for each word pair $(i,j)\in \mathcal{S}_{a_1}\times\mathcal{S}_{a_2}$, and for all \emph{synset} pairs $(\mathcal{S}_{a_1},\mathcal{S}_{a_2}) \in \mathcal{A}\times\mathcal{A}$.
    \item Let $\mathcal{C}$ be the union of all object \emph{synsets} annotated in an image. For each image in VisualGenome, $X_{ij}^{c}$ is incremented by $1$, for each word pair $(i,j)\in \mathcal{S}_{c_1}\times\mathcal{S}_{c_2}$, and for all \emph{synset} pairs $(\mathcal{S}_{c_1},\mathcal{S}_{c_2}) \in \mathcal{C}\times\mathcal{C}$.
    \item Let $\mathcal{H}$ be a set of object synsets annotated for an image in ImageNet and its ancestors in WordNet. For each each image in ImageNet, $X_{ij}^{oh}$ is incremented by $1$, for each word pair $(i,j)\in \mathcal{S}_{h_1}\times\mathcal{S}_{h_2}$, and for all \emph{synset} pairs $(\mathcal{S}_{h_1},\mathcal{S}_{h_2}) \in \mathcal{H}\times\mathcal{H}$.
\end{itemize}

%% file: sections/analysis.tex
\section{Experiments}~\label{sec:analysis}
We analyze ViCo embeddings with respect to the following  properties: 
(1) Does unsupervised clustering result in a natural grouping of words by visual concepts? (Sec.~\ref{sec:unsupervised}); 
(2) Do the word embeddings enable transfer of visual learning (\eg, visual recognition) to classes not seen during training? (Sec.~\ref{sec:zero-shot});
(3) How well do the embeddings perform on downstream applications? (Sec.~\ref{sec:downstream_tasks});
(4) Does the embedding space show word arithmetic properties ($land-car+aeroplane=sky$)? (Sec.~\ref{sec:analogies}).\footnote{We also perform a \emph{supervised partitioning} analysis which is included in the supplementary material. The results show that a supervised classification algorithm partitions words into visual categories more easily in the ViCo embedding space than in the GloVe or random vector space.}


\textbf{Data for clustering analysis.}
To answer (1) we manually annotate $495$ frequent words in VisualGenome with $13$ coarse (see legend in the t-SNE plots in Fig.~\ref{fig:analysis}) and $65$ fine categories (see appendix for the list of categories).


\textbf{Data for zero-shot-like analysis.} 
To answer (2), we use CIFAR-100~\cite{krizhevsky2009cifar}. We generate 4 splits of the 100 categories into disjoint Seen (categories used for training visual classifiers) and Unseen (categories used for evaluation)  sets. We use the following scheme for splitting: The list of 5  sub-categories in each of the 20 coarse categories (provided by CIFAR) is sorted alphabetically and the first $k$ categories are added to Seen and the remaining to Unseen for $k \in \{1,2,3,4\}$.

\input{figures/analysis_v2.tex}

\subsection{Unsupervised Clustering Analysis}~\label{sec:unsupervised}
The main benefit of word vectors over one-hot  or random vectors is the meaningful structure captured in the embedding space: words that are closer in the embedding space are semantically similar. We hypothesize that  ViCo  represents similarities and differences between visual categories that are missing from GloVe. 

Qualitative evidence to support this hypothesis can be found in t-SNE plots shown in Fig.~\ref{fig:analysis}, where concatenation of GloVe and ViCo embeddings leads to tighter, more homogenous clusters of the 13 coarse categories than GloVe. 

To test the hypothesis quantitatively, we cluster word embeddings with agglomerative clustering (cosine affinity and average linkage) and compare to the coarse and fine ground truth annotations using \emph{V-Measure} which is the harmonic mean of \emph{Homogeneity} and \emph{Completeness} scores. \emph{Homogeneity} is a measure of cluster purity, assessing whether all points in the same cluster have the same ground truth label. \emph{Completeness} measures whether all points with the same label belong to the same cluster\footnote{Analysis with other metrics and methods yields similar conclusions and is included in the supplementary material.}.

Plots (c,d) in Fig.~\ref{fig:analysis} compare random vectors, GloVe, variants of ViCo and their combinations (concatenation) for different number of clusters using V-Measure. Average performance across different cluster numbers is shown in Tab.~\ref{tab:analysis_baselines} and Tab.~\ref{tab:analysis_variants}. The main conclusions are as follows:

\textbf{ViCo clusters better than other embeddings.} Tab.~\ref{tab:analysis_baselines} shows that \emph{ViCo} alone outperforms
\emph{GloVe}, \emph{random}, and \emph{vis-w2v} based embeddings. \emph{GloVe+ViCo} improves performance further, especially for coarse categories.
\textbf{WordNet is not the sole contributor to strong performance of ViCo.} To verify that ViCo's gains are not simply due to the hierarchical nature of WordNet, we evaluate a version of ViCo trained on co-occurrences computed without using WordNet, \ie, using raw \emph{word} annotations in VisualGenome instead of \emph{synset} annotations and without Object-Hypernym co-occurrences. Tab.~\ref{tab:analysis_baselines} shows that \emph{GloVe+ViCo(linear,100,w/o WordNet)} outperforms \emph{GloVe} for both coarse and fine categories on both metrics. 




\textbf{ViCo outperforms existing visual word embeddings.} Tab.~\ref{tab:analysis_baselines} evaluates performance of existing visual word embeddings which are learned from abstract scenes~\cite{kottur2016visw2v}. \emph{wiki} and \emph{coco} are different versions of \emph{vis-w2v} depending on the dataset (Wikipedia or MS-COCO~\cite{lin2014mscoco,chen2015cococaptions}) used for training word2vec for initialization. After initialization, both models are trained on an abstract scenes (clipart images) dataset~\cite{zitnick2013visualabstraction}. \emph{ViCo(linear,100)} outperforms both of these embeddings. \emph{GloVe+vis-w2v-wiki} performs similarly to \emph{GloVe} and \emph{GloVe+vis-w2v-wiki-coco} performs only slightly better than \emph{GloVe}, showing that the majority of the information captured by \emph{vis-w2v} may already be present in \emph{GloVe}.

\textbf{Learned embeddings significantly outperform random vectors.} Tab.~\ref{tab:analysis_baselines} shows that random vectors perform poorly in comparison to learned embeddings. \emph{GloVe+random} performs similarly to \emph{GloVe} or worse. This implies that gains of \emph{GloVe+ViCo} over \emph{GloVe} are not just an artifact of increased dimensionality.

\textbf{\emph{Linear} achieves similar performance as \emph{Select} with fewer dimensions.} Tab.~\ref{tab:analysis_variants} illustrates the ability of the multi-task formulation to learn a more compact representatio than \emph{select} (concatenating embeddings learned from each co-occurrence type separately) without sacrificing performance. $50$, $100$, and $200$ dimensional ViCo embeddings learned with linear transformations, all achieve performance similar to \emph{select}.

\subsection{Zero-Shot-like Analysis}~\label{sec:zero-shot}
The ability of word embeddings to capture relations between visual categories  enables 
to generalize visual models trained on limited visual categories to larger sets unseen during training. 
To assess this ability, we evaluate embeddings on their zero-shot-like object classification performance using the CIFAR-100 dataset. Note that our \emph{zero-shot-like} setup is slightly different from a typical zero-shot setup because even though the visual classifier is not trained on unseen class images in CIFAR, annotations associated with images of unseen categories in VisualGenome or ImageNet may be used to compute word co-occurrences while learning word embeddings. \\

\input{tables/analysis_unsup_baselines.tex}
\input{tables/analysis_unsup_vico_variants.tex}

\noindent\textbf{Model.} Let $f(I) \in \mathbb{R}^n$ be the features extracted from image $I$ using a CNN and let $w_c \in \mathbb{R}^m$ denote the word embedding for class $c \in \mathcal{C}$. Let $g: \mathbb{R}^{m} \rightarrow \mathbb{R}^n$ denote a function that projects word embeddings into the space of image features. We define the score $s_c(I)$ for class $c$ as $\text{cosine}(f(I),g(w_c))$, 
where $\text{cosine}(\cdot)$ is the cosine similarity. The class probabilities are defined as  
\begin{equation}
    p_c(I) = \frac{\exp( s_c(I)/\epsilon)}
    {\sum_{c' \in \mathcal{C}} \exp( s_{c'}(I)/\epsilon)},
\end{equation}
where $\epsilon$ is a learnable temperature parameter. In our experiments, $f(I)$ is a $64$-dimensional feature vector produced by the last linear layer of a 34-layer ResNet (modified to accept $32 \times 32$ CIFAR images) and $g$ is a linear transformation. \\

\noindent
\textbf{Learning.} The model (parameters of $f$, $g$, and $\epsilon$) is trained on images from the set of seen classes $\mathcal{S} \subset \mathcal{C}$. We use the Adam~\cite{kingma2014adam} optimizer with a learning rate of $0.01$. The model is trained with a batch size of $0.01$ for $50$ epochs. \\

\noindent
\textbf{Model Selection and Evaluation.} The best model (among iteration checkpoints) is selected based on seen class accuracy (classifying only among classes in $\mathcal{S}$) on the test set. The selected model is evaluated on unseen category ($\;\mathcal{U} = \mathcal{C} \setminus \mathcal{S}$) prediction accuracy  computed on the test set. 

Fig.~\ref{fig:zero_shot} compares chance performance ($1/|\mathcal{U}|$), random vectors, \emph{GloVe}, and \emph{GloVe+ViCo} on four seen/unseen splits. We show mean and standard deviation computed across four runs ($7\times4\times4=112$ models trained in all). The key conclusions are as follows:

\textbf{ViCo generalizes to unseen classes better than GloVe.} ViCo based embeddings, especially $200$-dim.\ select and linear variants show healthy gains over \emph{GloVe}. Note that this is not just due to higher dimensions of the embeddings since \emph{GloVe+random(200)} performs worse than \emph{GloVe}.

\textbf{Learned embeddings significantly outperform random vectors.} Random vectors alone achieve close to chance performance, while concatenating random vectors to \emph{GloVe} degrades performance. 

\textbf{Select performs better than Linear.} Compression to $100$-dimensional embeddings using linear transformation shows a more noticeable drop in performance as compared to the \emph{select} setting. However, \emph{GloVe+ViCo(linear,100)} still outperforms \emph{GloVe} in 3 out of 4 splits. 


\input{figures/zero_shot.tex}
\input{tables/tasks_baslines.tex}

%% file: figures/analysis_v2.tex
\begin{figure*}[t]
\vspace{-0.75em}
\begin{center}
\includegraphics[width=\linewidth]{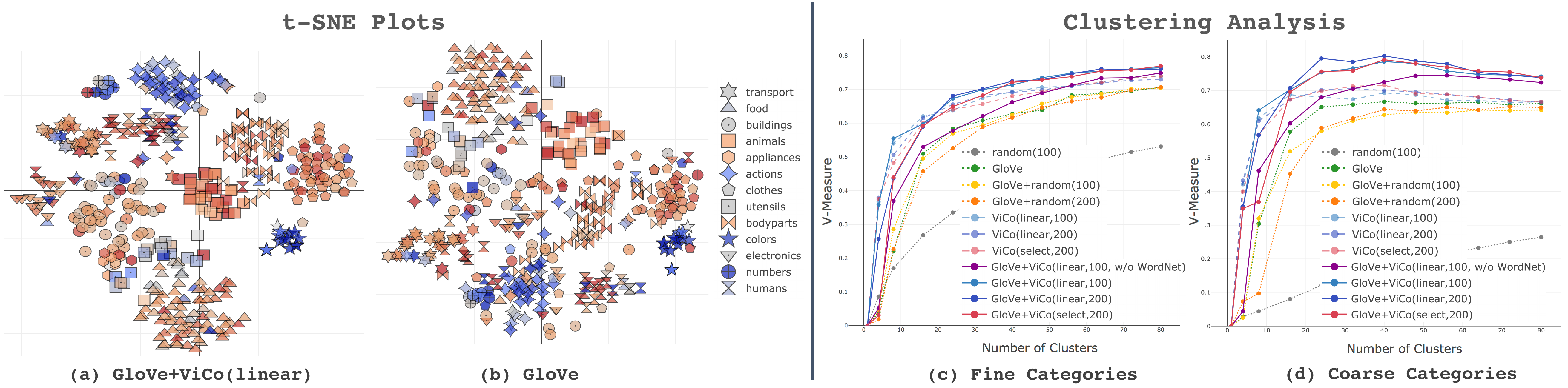}
\end{center}
\vspace{-0.75em}
\caption{\textbf{Unsupervised Clustering Analysis.} (a,b) \textbf{Qualitative evaluation with t-SNE:} Plots show that ViCo augmented GloVe results in tighter, more homogenous clusters than GloVe. Marker shape encodes the annotated coarse category and color denotes if the word is used more frequently as an {\color{brickred} object} or an {\color{darkblue} attribute}; (c,d) \textbf{Quantitative evaluation:} Plots show clustering performance of different embeddings measured through V-Measure at different number of clusters. All ViCo based embeddings outperform GloVe for both fine and coarse annotations (Sec.~\ref{sec:unsupervised}). See Tab.~\ref{tab:analysis_baselines} and Tab.~\ref{tab:analysis_variants} for average performance across cluster numbers. Best viewed in color on a screen.}
\label{fig:analysis}
\vspace{-1em}
\end{figure*}


%% file: tables/analysis_unsup_baselines.tex
\renewcommand{\arraystretch}{1.1}
\setlength{\tabcolsep}{3pt}
\begin{table}[t]
\centering
\footnotesize
\vspace{-0.4cm}
\begin{tabular}{l|c|c|c}
\textbf{Embeddings} & \textbf{Dim.} & \textbf{Fine} & \textbf{Coarse}\\
\hline
random(100) & 100 & 0.34 & 0.15  \\
GloVe & 300 & 0.50 & 0.52  \\
GloVe+random(100) & 300+100 & 0.50 &  0.49  \\
\hline
vis-w2v-wiki~\cite{kottur2016visw2v}       & 200     & 0.41 & 0.43  \\
vis-w2v-coco~\cite{kottur2016visw2v}       & 200 & 0.45 & 0.4   \\
GloVe+vis-w2v-wiki & 300+200 & 0.5  & 0.52  \\
GloVe+vis-w2v-coco & 300+200 & 0.52 & 0.55 \\
\hline
ViCo(linear,100) & 100 & \sbest{0.60} & \sbest{0.59} \\
GloVe+ViCo(linear,100) & 300+100 & \best{0.61} & \best{0.65} \\
GloVe+ViCo(linear,100, w/o WN) & 300+100 & 0.54 & 0.58 \\
\end{tabular}
\vspace{0.5em}
\caption{\textbf{Comparing ViCo to other embeddings.} All ViCo based embeddings 
outperform GloVe and random vectors. \emph{ViCo(linear,100)} also outperforms \emph{vis-w2v}. \emph{GloVe+vis-w2v} performs similarly to \emph{GloVe} while \emph{GloVe+ViCo} outperforms both \emph{GloVe} and ViCo. Using WordNet yields healthy performance gains but is not the only contributor to performance since \emph{GloVe+ViCo(linear,100, w/o WN)} also outperforms \emph{GloVe}. \best{Best} and \sbest{second best} numbers are highlighted in each column.}
\label{tab:analysis_baselines}
\end{table}
\renewcommand{\arraystretch}{1}

%% file: tables/analysis_unsup_vico_variants.tex
\renewcommand{\arraystretch}{1.1}
\setlength{\tabcolsep}{4pt}
\begin{table}[t]
\centering
\footnotesize
\vspace{-1em}
\begin{tabular}{l|c|c|c}
\textbf{Embeddings} & \textbf{Dim.} & \textbf{Fine} & \textbf{Coarse} \\
\hline
ViCo(linear,50)    & 50      & 0.57 & 0.56 \\
ViCo(linear,100)   & 100     & \sbest{0.60} & 0.59 \\
ViCo(linear,200)   & 200     & 0.59 & 0.60  \\
ViCo(select,200)   & 200     & 0.59 & 0.60  \\
\hline
GloVe & 300 & 0.50 & 0.52 \\
GloVe+ViCo(linear,50) & 300+50 & 0.60 & \best{0.66} \\
GloVe+ViCo(linear,100) & 300+100 & \best{0.61} & \sbest{0.65} \\
GloVe+ViCo(linear,200) & 300+200 & \sbest{0.60} & \sbest{0.65} \\
GloVe+ViCo(select,200) & 300+200 & 0.57 & 0.63 \\
\end{tabular}
\vspace{0.5em}
\caption{\textbf{Effect of transformations on clustering performance.} The table compares average performance across number of clusters. The \emph{linear} variants achieve performance similar to \emph{select} with fewer dimensions. In fact, when used in combination with GloVe, \emph{linear} variants outperform \emph{select}. \best{Best} and \sbest{second best} numbers are highlighted in each column.}
\label{tab:analysis_variants}
\vspace{-1em}
\end{table}
\renewcommand{\arraystretch}{1}

%% file: figures/zero_shot.tex
\begin{figure}[t]
\vspace{-0.4cm}
\begin{center}
\includegraphics[width=\linewidth]{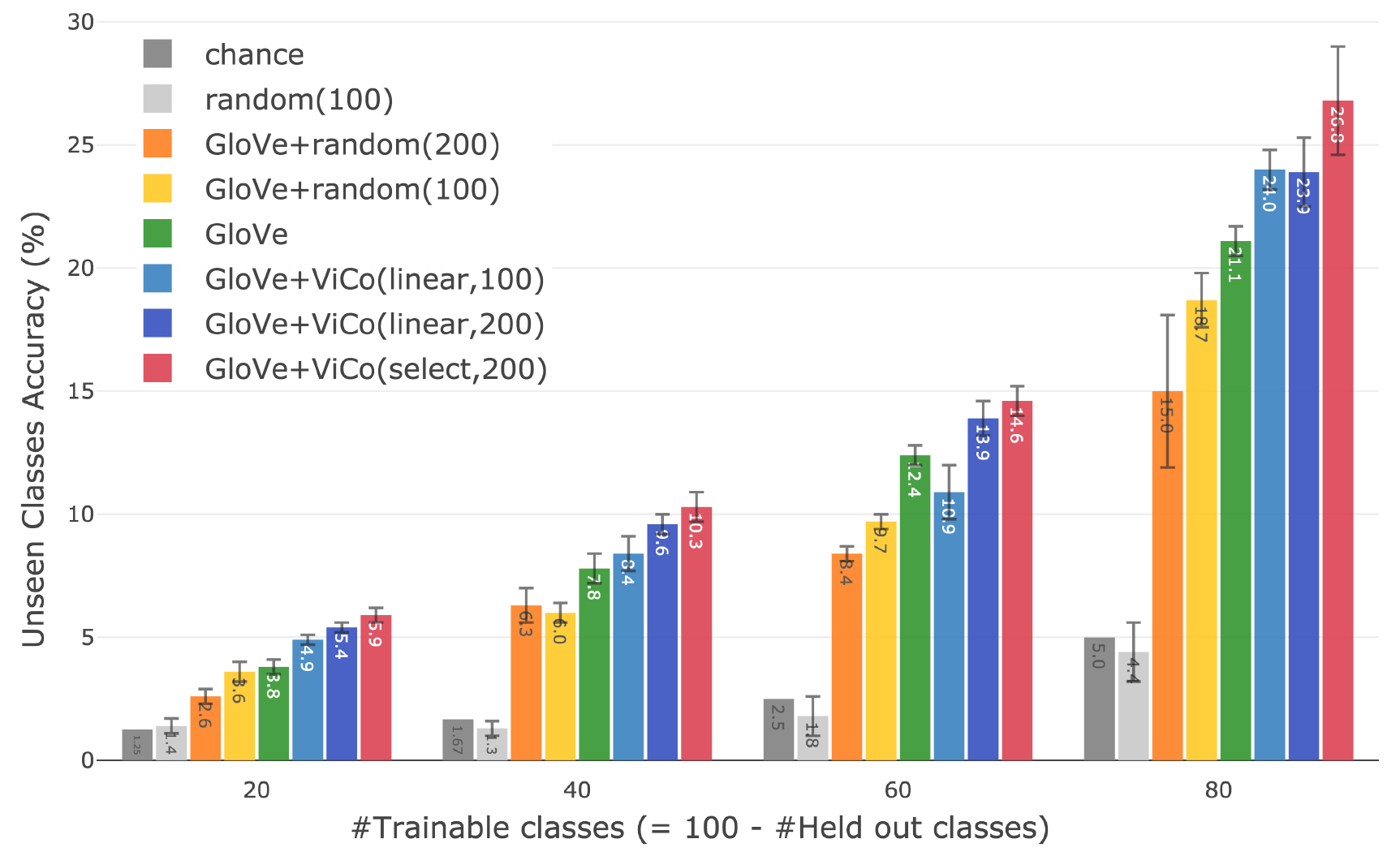}
\end{center}
\vspace{-0.75em}
\caption{\textbf{Zero-Shot Analysis.} The histogram compares the transfer learning ability of a simple word embedding based object classification model. The $x$-axis denotes the number of CIFAR-100 classes ($m$) used during training. During test, we evaluate the classifier on its ability to correctly classify among the remaining ($100-m$) unseen classes. Results show that \emph{GloVe+ViCo} leads to better transfer to unseen classes than GloVe alone (Sec.~\ref{sec:zero-shot}).}
\vspace{-1em}
\label{fig:zero_shot}
\end{figure}

%% file: tables/tasks_baslines.tex
\renewcommand{\arraystretch}{1.2}
\setlength{\tabcolsep}{2.7pt}
\begin{table*}[t]
\footnotesize
\centering
\vspace{-0.5cm}
\begin{tabular}{l|l|c|cc|cccc|ccc|cccc}
& & \textbf{\color{brickred}Discr. Attr.} & \multicolumn{2}{c|}{\textbf{\color{darkblue}Im-Cap Retrieval}} & \multicolumn{4}{c|}{\textbf{\color{darkblue}VQA}} & \multicolumn{3}{c|}{\textbf{\color{darkblue}Ref. Exp.}} & \multicolumn{4}{c}{\textbf{\color{darkblue}Image Captioning}}\\ 
& & Avg. F1 & \multicolumn{2}{c|}{Recall@1} & \multicolumn{4}{c|}{Accuracy} & \multicolumn{3}{c|}{Loc. Accuracy} & \multicolumn{4}{c}{Captioning Metrics}\\ 
\textbf{Embeddings}            & \textbf{Dim.} & $m\pm \sigma$                   & Im2Cap & Cap2Im & Overall & Y/N & Num. & Other & Val & TestA & TestB & B1 & B4 & C & S \\
\hline
random                 & 300           & 50.03 $\pm$ 2.26         & 43.1                   & 30.6            & 66.1             & 82.0            & 44.8            & 57.5           & 71.3                & 73.5           & 66.3           & \best{0.714}   & \best{0.296}   & \best{0.910}   & \best{0.170}          \\
GloVe                  & 300           & 63.85 $\pm$ 0.04         & \sbest{44.8}                   & 33.5            & \sbest{67.5}             & 83.8            & \sbest{46.5}    & \sbest{58.3}   & 72.2                & \sbest{75.3}   & 66.8           & 0.708          & 0.290          & 0.891          & 0.167          \\
GloVe + random         & 300+100       & \sbest{63.88 $\pm$ 0.03}         & 44.3                   & \best{34.4}    & \sbest{67.5}             & \sbest{84.1}            & 45.9            & 58.2           & \sbest{72.5}       & 75.1           & \best{67.5}    & 0.707          & 0.288          & 0.881          & 0.166          \\
GloVe + ViCo (linear)  & 300+100       & \best{64.46 $\pm$ 0.17}  & \best{46.3}           & \sbest{34.2}            & \best{67.7}      & \best{84.4}     & \best{46.6}     & \best{58.4}    & \best{72.7}         & \best{75.5}    & \best{67.5}    & \sbest{0.711}  & \sbest{0.291}  & \sbest{0.894}  & \sbest{0.168}  
\end{tabular}
\vspace{0.5em}
\caption{\textbf{Comparing ViCo to GloVe and random vectors.} \emph{GloVe+ViCo(linear)} outperforms \emph{GloVe} and \emph{GloVe+random} for all tasks and outperforms \emph{random} for all tasks except Image Captioning. While random vectors perform close to chance on the \textbf{\color{brickred}word-only} task, they compete tightly with learned embeddings on \textbf{\color{darkblue}{vision-language}} tasks. This suggests that vision-language models are relatively insensitive to the choice of word embeddings. \best{Best} and \sbest{second best} numbers in each column are highlighted.}
\label{tab:tasks_baselines}
\vspace{-0.5cm}
\end{table*}
\renewcommand{\arraystretch}{1}

%% file: sections/tasks.tex
\subsection{Downstream Task Evaluation}~\label{sec:downstream_tasks}
We now evaluate ViCo embeddings on a range of downstream tasks. Generally, we expect tasks requiring better word representations of objects and attributes to benefit from our embeddings. When using existing models, we initialize and freeze word embeddings so that performance changes are not due to fine-tuning embeddings of different dimensions. The rest of the model is left untouched except for the dimensions of the input layer where the size of the input features needs to match the embedding dimension. 

Tab.~\ref{tab:tasks_baselines} compares performance of embeddings on a word-only discriminative attributes task and 4 vision-language tasks. On all tasks \emph{GloVe+ViCo} outpeforms \emph{GloVe} and \emph{GloVe+random}. Unlike the word-only task which depends solely on word representations, vision-language tasks are less sensitive to word embeddings, with performance of random embeddings approaching learned embeddings \footnote{See supplementary material for our hypothesis and test for why random vectors work well for vision-language tasks.}. 

\textbf{Discriminative Attributes}~\cite{krebs2018discrattr} is one of the SemEval 2018 challenges. The task requires to identify whether an attribute word discriminates between two concept words. For example, the word ``red'' is a discriminative attribute for word pair (``apple'', ``banana'') but not for (``apple'', ``cherry''). Samples are presented as tuples of attribute and concept words and the model makes a binary prediction. Performance is evaluated using class averaged F1 scores. 

Let $w_1$, $w_2$, and $a$ be the word embeddings (GloVe or ViCo) for the two concept words and the attribute word. We compute the scores $s_g$ and $s_v$ for GloVe and ViCo using function $s(a,w_1,w_2) = \text{cosine}(a,w_1) - \text{cosine}(a,w_2)$, where $\text{cosine}(\cdot)$ is the cosine similarity. We then learn a linear SVM over $s_g$ for the \emph{GloVe} only model and over $s_g$ and $s_v$ for the \emph{GloVe+ViCo} model.



\textbf{Caption-Image Retrieval} is a classic vision-language task requiring a model to retrieve images given a caption or vice versa. We use the open source VSE++~\cite{faghri2017vse++} implementation which learns a joint embedding of images and captions using a \emph{Max of Hinges} loss that encourages attending to hard negatives and is geared towards improving top-1 Recall. We evaluate the model using Recall@1 on MS-COCO. 


\textbf{Visual Question Answering}~\cite{antol2015vqa,goyal2017vqa2} systems are required to answer questions about an image. We compare the performance of embeddings using Pythia~\cite{jiang2018pythia,jiang2018pythiacode} which uses bottom-up top-down attention for computing a question-relevant image representation. Image features are then fused with a question representation using a GRU operating on word embeddings and fed into an answer classifier. Performance is evaluated using overall and by-question-type accuracy on the test-dev split of the VQA v2.0 dataset. 


\textbf{Referring Expression Comprehension}  consists of localizing an image region based on a natural language description. We use the open source implementation of MAttNet~\cite{yu2018mattnet} to compare localization accuracy with different embeddings on the RefCOCO+ dataset using the UNC split. MAttNet uses an attention mechanism to parse the referring expression into phrases that inform the subject's appearance, location, and relationship to other objects. These phrases are processed by corresponding specialized localization modules. The final region scores are 
 a linear combination of module scores using predicted weights. 

\textbf{Image Captioning} involves generating a caption given an image. We use the Show and Tell model of Vinyals~\etal~\cite{vinyals2015showtell} which feeds CNN extracted image features into an LSTM followed by beam search to sample  captions. We report BLEU1 (B1), BLEU4 (B4), CIDEr (C), and SPICE (S) metrics~\cite{papineni2002bleu,vedantam2015cider,anderson2016spice} on the MS-COCO test set.

\subsection{Exploring Embedding Space Structure}~\label{sec:analogies}
Previous work~\cite{mikolov2013w2vnaacl} has demonstrated linguistic regularities in word embedding spaces through analogy tasks solved using simple vector arithmetics. Fig.~\ref{tab:analogies} shows qualitatively that ViCo embeddings possess similar properties, capturing relations between visual concepts well. 

\input{tables/analogies.tex}

\section{Conclusion}
\vspace{-0.2em}
This work shows that in addition to textual co-occurrences, visual co-occurrences are a surprisingly effective source of information for learning word representations. The resulting embeddings outperform text-only embeddings on unsupervised clustering, supervised partitioning, zero-shot generalization, and various supervised downstream tasks. We also develop a multi-task extension of \emph{GloVe}'s log-bilinear model to learn a compact shared embedding from multiple types of co-occurrences. Type-specific embedding spaces learned as part of the model help provide a richer sense of relatedness between words. \\

\vspace{-0.4em}
\noindent\textbf{Acknowledgments:} Supported in part by  NSF 1718221, ONR MURI N00014-16-1-2007, Samsung, and 3M.

%% file: tables/analogies.tex
\renewcommand{\arraystretch}{1.1}
\setlength{\tabcolsep}{2pt}
\vspace{-0.75em}
\begin{table}[h]
\centering
\scriptsize
\begin{tabular}{l|c|c|c}
\textbf{Analogy} & \textbf{Answer Candidates} & \textbf{GloVe} & \textbf{ViCo}\\
\hline
car:land::aeroplane:? & ocean, sky, road, railway & ocean & \textbf{sky}\\
clock:circle::tv:? & triangle, square, octagon, round & triangle & \textbf{square} \\
park:bench::church:? & door, sofa, cabinet, pew & door & \textbf{pew} \\
sheep:fur::person:? & hair, horn, coat, tail & coat & \textbf{hair} \\
monkey:zoo::cat:? & park, house, church, forest & park & \textbf{house} \\
leg:trouser::wrist:? & watch, shoe, tie, bandana & bandana & \textbf{watch} \\
yellow:banana::red:? & strawberry, lemon, mango, orange & mango & \textbf{strawberry} \\
rice:white::spinach:? & blue, green, red, yellow & blue & \textbf{green} \\
train:railway::car:? & land, desert, ocean, sky & land & \textbf{land} \\
can:metallic::bottle:? & wood, glass, cloth, paper & glass & \textbf{glass} \\
\hline
man:king::woman:? & queen, girl, female, adult & \textbf{queen} & girl \\
can:metallic::bottle:? & wood, plastic, cloth, paper & \textbf{plastic} & wood \\
train:railway::car:? & road, desert, ocean, sky & \textbf{road} & ocean
\end{tabular}
\vspace{0.5em}
\caption{\textbf{Answering Analogy Questions.} Out of 30 analogy pairings tested, we found both GloVe and ViCo to be correct 19 times, only ViCo was correct 8 times, and only Glove was correct 3 times. Correct answers are \textbf{highlighted}.}
\label{tab:analogies}
\vspace{-2em}
\end{table}
\renewcommand{\arraystretch}{1.0}